\documentclass[review]{elsarticle}

\usepackage[margin=2.5cm]{geometry}

\usepackage{lineno,hyperref}
\usepackage[utf8]{inputenc}
\usepackage{subfig}
\usepackage{float}
\usepackage{svg}

\usepackage{array}
\usepackage{mathtools}
\newcolumntype{C}[1]{>{\centering\arraybackslash}p{#1}}

\modulolinenumbers[5]

\journal{Expert Systems With Applications}

\begin{document}

\begin{frontmatter}

%\title{Automatic Diagnosis of the Ullrich Congenital Muscular Dystrophy \\ using Convolutional Neural Networks}
\title{A Convolutional Neural Network for the Automatic Diagnosis of \\ Collagen VI related Muscular Dystrophies}

\cortext[cor1]{Corresponding authors.}

\author[IRI]{Adrián Bazaga}
\ead{arodriguezb@iri.upc.edu}

\author[JDD1]{Mònica Roldán}
\ead{mroldanm@sjdhospitalbarcelona.org}

\author[JDD2]{Carmen Badosa}
\ead{mcbadosa@fsjd.org}

\author[JDD2]{Cecilia Jiménez-Mallebrera\corref{cor1}}
\ead{cjimenezm@fsjd.org}

\author[IRI]{Josep M. Porta\corref{cor1}}
\ead{porta@iri.upc.edu}

\address[IRI]{Institut de Robòtica i Informàtica Industrial, UPC-CSIC, 08028 Barcelona, Spain}

\address[JDD1]{Unitat de Microscòpia Confocal, Servei d'Anatomia Patològica, Institut Pediàtric de Malalties Rares, Hospital de Sant Joan de Déu, 08950 Barcelona, Spain}

\address[JDD2]{Hospital de Sant Joan de Déu, 08950 Barcelona, Spain}

%-----------------------------------------------------------------------------------------------------

\begin{abstract}
The development of machine learning systems for the diagnosis of rare diseases is challenging mainly due the lack of data to study them. Despite this challenge, this paper proposes a system for the Computer Aided Diagnosis (CAD) of low-prevalence, congenital muscular dystrophies from confocal microscopy images. The proposed CAD system relies on a Convolutional Neural Network (CNN) which performs an independent classification for non-overlapping patches tiling the input image, and generates an overall decision summarizing the individual decisions for the patches on the query image. This decision scheme points to the possibly problematic areas in the input images and provides a global quantitative evaluation of the state of the patients, which is fundamental for diagnosis and to monitor the efficiency of therapies.
\end{abstract}

%-----------------------------------------------------------------------------------------------------
\begin{keyword}
Convolutional neural networks \sep Deep learning \sep Classification \sep Computer aided diagnosis \sep Confocal microscopy images \sep Collagen~VI related muscular dystrophies. 
% Limited to 6 keywords, removed: Expert systems
\end{keyword}

\end{frontmatter}

%\linenumbers

%-----------------------------------------------------------------------------------------------------
%-----------------------------------------------------------------------------------------------------
%-----------------------------------------------------------------------------------------------------

\section{Introduction}

Deficiencies in the structure of collagen~VI are a common cause of neuromuscular diseases with manifestations ranging from the Bethlem myopathy to the severe Ullrich congenital muscular dystrophy. The symptoms include proximal and axial muscle weakness, distal hyperlaxity, joint contractures, and critical respiratory insufficiency, which requires assisted ventilation and results in a reduced live expectancy. Moreover, the skin and other connective tissues where collagen VI is abundant are also affected~\cite{Nadeau:2009,Lamand:2018}. The collagen~VI structural defects are related to mutations of three main genes (COL6A1, COL6A2, and COL6A3, OMIM 254090 and 158810). Thus, the new advances in genome editing tools open the possibility to successfully treat these neuromuscular diseases  for the first time. This opportunity, though, comes with important challenges. Beyond the challenges of gene editing, this paper focuses on the challenges arising when trying to formally evaluate the efficiency of the therapeutic approaches in the recovery of the collagen~VI microfibrillar network. 

\begin{figure}
  \centering
  \includegraphics[height=155px]{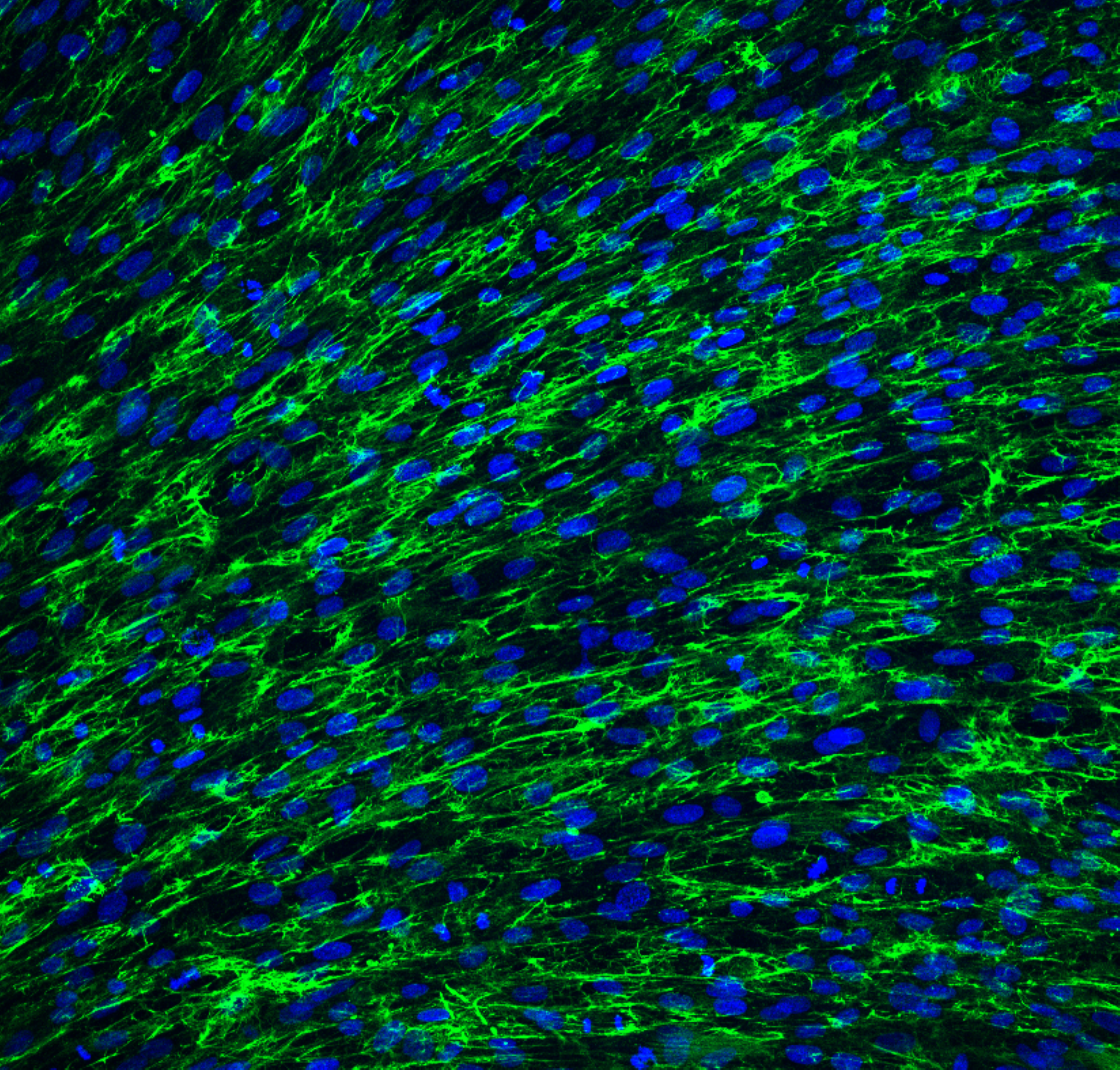}
  \includegraphics[height=155px]{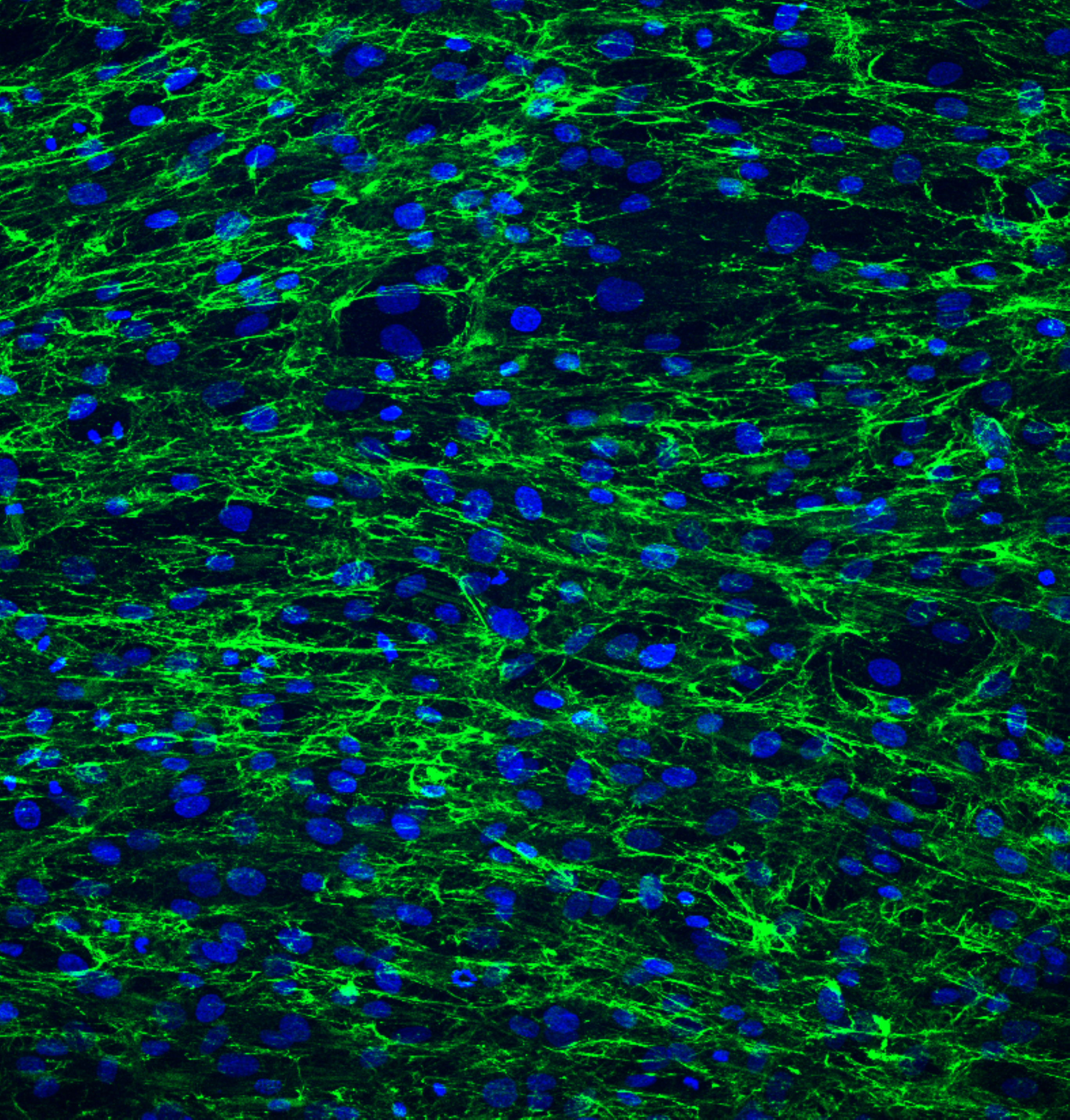}
  \includegraphics[height=155px]{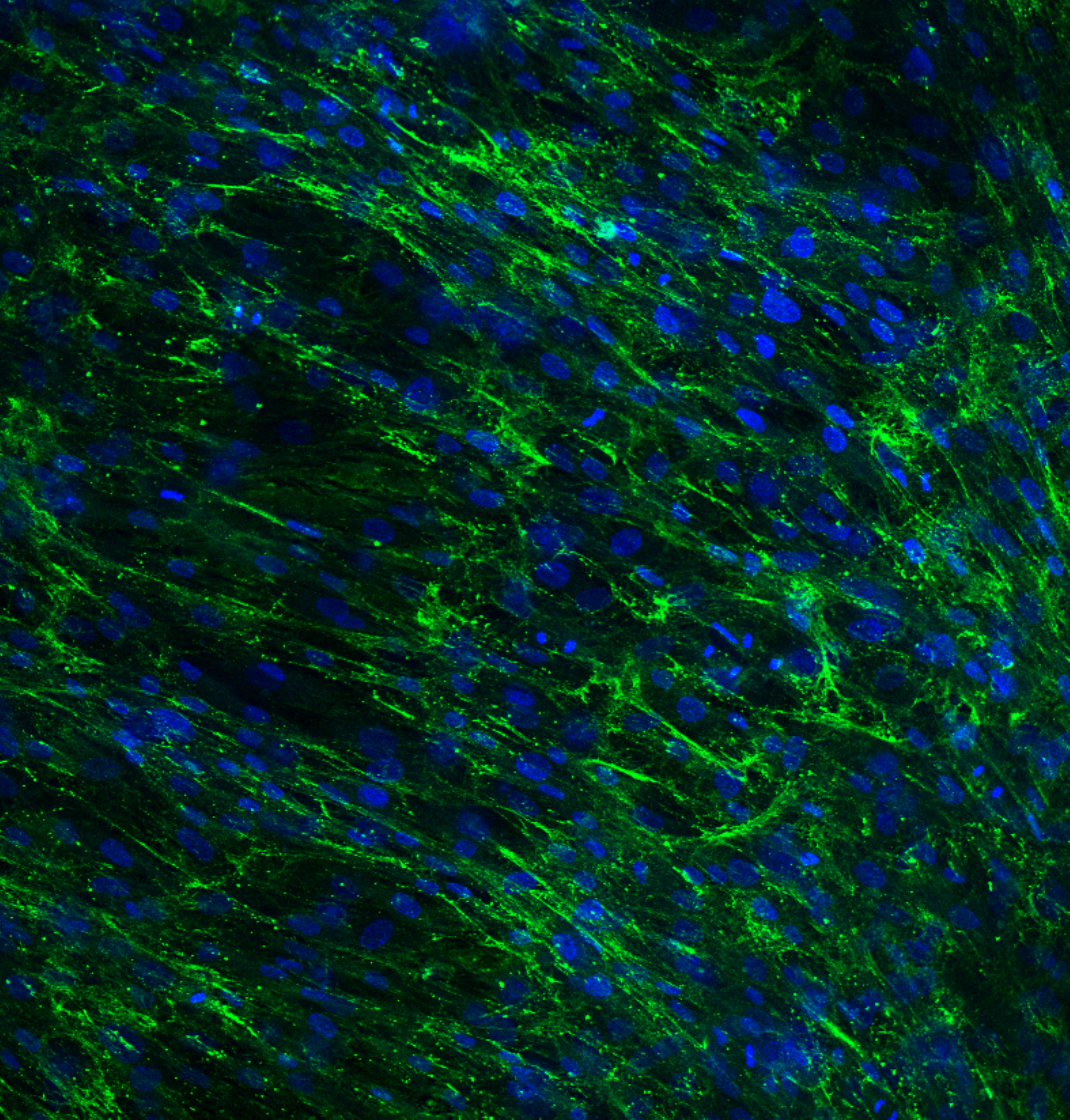}
  \caption{Confocal microscopy images of fibroblast cultures. Left: A control sample. Center: Sample from a patient with the Bethlem myopathy. Right: Sample from a patient of the Ullrich muscular dystrophy. In the three images, the network of collagen is shown in green and the fibroblast nuclei in blue.}
  \label{biopsy}
\end{figure}

Despite the implementation of next generation of gene sequencing, the diagnosis of rare diseases remains challenging. This is particularly true for diseases caused by dominant mutations where there is not a complete absence of a main protein and when the effect of a genetic variant on the protein structure may not be evident. Thus, prior to genetic analysis or to demonstrate the pathogenicity of novel mutations, the standard technique for the diagnosis of collagen VI related dystrophies is the analysis of images of fibroblast cultures~\cite{Jimenez-Mallebrera:2006} (see Fig.~\ref{biopsy}). Several aspects of the images such as the orientation of the collagen fibers, the regularity of the collagen network, and the arrangement of cells in such network are taken into account by the specialists to identify the patients. However, this evaluation is only qualitative and the regulatory agencies would not approve any treatment without an objective proof of its effectiveness~\cite{Anthony:2014}. Thus, there is an acute need of accurate methodologies to quantitatively monitor the effects of any possible new therapy.

This paper proposes a computer aided diagnosis (CAD) system for dystrophies caused by defects in the structure of collagen~VI. The system uses a Convolutional Neural Network (CNN) to classify fibroblast culture images from possible patients. The CNN is trained from cases labeled by specialists but, as in the case of all rare diseases, there is a severe lack of training data. To address this issue, this paper uses a patch-based data-augmentation technique similar to the one proposed in~\cite{Krizhevsky:2012}.

This paper is structured as follows: Section~\ref{sec:related} frames the contribution of this paper in the context of CAD systems using medical images and Section~\ref{sec:preliminaries} introduces the basic elements defining a CNN. Then,
Section~\ref{sec:system} describes in detail the proposed system, Section~\ref{sec:setup} discusses the image acquisition and the
data augmentation scheme, and Section~\ref{sec:results} presents the results. Finally, Section~\ref{sec:conclusions} concludes the paper and points to issues deserving further attention.

%-----------------------------------------------------------------------------------------------------
%-----------------------------------------------------------------------------------------------------
%-----------------------------------------------------------------------------------------------------

\section{Related work} \label{sec:related}

The automatic analysis of medical images is typically divided in two steps. In the first one, relevant features are extracted from the image. In the second step, those features are used to evaluate a particular function that in our case is the classification between control and patient images.

Several methodologies have been proposed to identify relevant features in medical images. For instance, in~\cite{Schilham:2006} a multi-scale Gaussian filterbank is used to find perturbations in chest X-ray images. However, in our case, the disease causes a global disorganization of the collagen network, rather than local perturbations on it. In~\cite{Adankon:2012} the authors use a principal component analysis technique to extract relevant features from an X-ray image. These features, however are linearly related to the input image and such a simple transformation is not likely to provide relevant features in our case. In other approaches, Localy Binary Patterns (LBP)~\cite{Ahonen:2006}, completed LBP~\cite{Guo:2010}, gray-level co-occurence matrices~\cite{Soh:1999}, or Artificial Neural Networks (ANNs)~\cite{Khatami:2017} have been also used to obtain features, all on input images significantly different from the ones available in our case. In works more related to the problem considered in this paper, some systems have been proposed to extract features from collagen images, such as the distribution of the fibers orientation~\cite{Sun:2015}. However, in our case, this distribution is only weakly related with the diagnostic, since some patients present coherently-oriented collagen networks (see Fig.~\ref{biopsy}-center). The same weak relation occurs with other possible features such as the average intensity of the collagen network, or the presence of collagen precipitates in the images. Thus, in the addressed problem, the selection of relevant features for the diagnostic is a challenge on itself.

Nevertheless, assuming that relevant features can be somehow identified, a classical approach would apply a classification method relying on them. Since there is not a single method adequate for all problems, different classifiers have been used on features extracted from medical images, including naive Bayes classifiers for the detection of brain tumors~\cite{Subashini:2016}, support vector machines for the classification of pigmented skin lesions~\cite{Celebi:2007} and breast cancer~\cite{Nahar:2012}, random forests for the diagnosis of tuberculosis~\cite{Shabut:2018}, non-parametric approaches to identify patients with diabetic retinopathy pathology~\cite{Venkatesan:2015}, and Convolutional Neural Networks (CNNs) for the detection of melanoma or glaucoma~\cite{Menegola:2016,Vinicius:2018}. In \cite{Sudharshan:2019} some of these approaches are compared on histopathological breast cancer images, which are similar to the images considered in this paper, and they conclude that non-parametric approaches and CNNs provide the best results. CNNs are an evolution of the classical ANNs. However, there exists two main differences between ANNs and CNNs. First, CNNs are invariant to translations of the features in the image due to shared parameters between the image convolutions. Second, new types of layers appear in the CNNs such as the so called max-pooling layers, which enlarge the receptive fields of the convolutions and also contribute to the translation invariance. Such improvements, together with the availability of specialized hardware and software for training CNNs have established them as the {\em de-facto} standard tool for image processing, provided that enough training data is available. An additional advantage of the CNNs is that they can directly operate on images, taking care of both identifying relevant features and performing the classification relying on them~\cite{Lecun:1998,Simonyan:2014}. This is the fundamental reason to apply them to the problem addressed in this paper.

%-----------------------------------------------------------------------------------------------------
%-----------------------------------------------------------------------------------------------------
%-----------------------------------------------------------------------------------------------------

\section{Preliminaries on CNNs} \label{sec:preliminaries}

A CNN is composed by a set of layers of simple computational elements, called neurons, connected to elements in the previous layers. The layers are evaluated in sequence, from the input to the output. The computation elements are pre-defined and fixed and, thus, the parameters encoding the learned function  are the weights associated with the connections in the network.

Three different kinds of layers are typically present in a CNN: convolutional layers, pooling layers, and fully connected layers \cite{Goodfellow:2016}.
Convolutional layers extract features applying a convolutional kernel all over the input image. Formally, each kernel is computed as
\begin{equation}
y=\sum_{n=1}^{N} I_n k_{n},
\label{convlayer1}
\end{equation}
where $I_n$ is a pixel in the image, $k_n$ is $n$-th kernel weight, and $N$ is the kernel size, i.e., the amount of pixels in the image affected by the kernel. Convolutions are typically applied on squared areas around selected pixels in the input image and, thus, the result of this operation, is a new image, $Y$, which is called a feature map. Often, the output of a kernel is passed through an activation function. For instance, in our approach, each kernel is followed by a rectified linear unit (ReLU) activation function, 
\begin{equation}
        f(x)=
        \left\{ \begin{array}{ll}
            0 & x < 0 \\
            x & x \geq 0
        \end{array} \right..
\label{reluformula}
\end{equation}
which eliminates negative inputs, introducing non-linearities in the CNN with a low computational cost.
The horizontal/vertical distance (in pixels) between the centers of two consecutive kernels is known as the stride of the kernel. Strides larger than one are typically used to downsample a feature map. 
Alternatively, the pooling layers down-sample the input by summarizing a patch in the image with a single value. For instance, in our work we use the max-pooling operation, where the output is the maximum of the inputs. 

Finally, the fully-connected layers are typically placed at the end of the network and they provide the classification decision by applying an activation function. In our case, we use the sigmoid activation function
\begin{equation}
\sigma(x)=\frac{1}{1+e^{-x}}, \label{sigmoidformula}
\end{equation}
where $x$ is a weighted sum of inputs and where the output is a number between 0 and 1.
%, with values close to 0 and to 1 for large negative and positive inputs, respectively. 

\begin{figure}[t]
  \centering
  \includegraphics[scale=0.5]{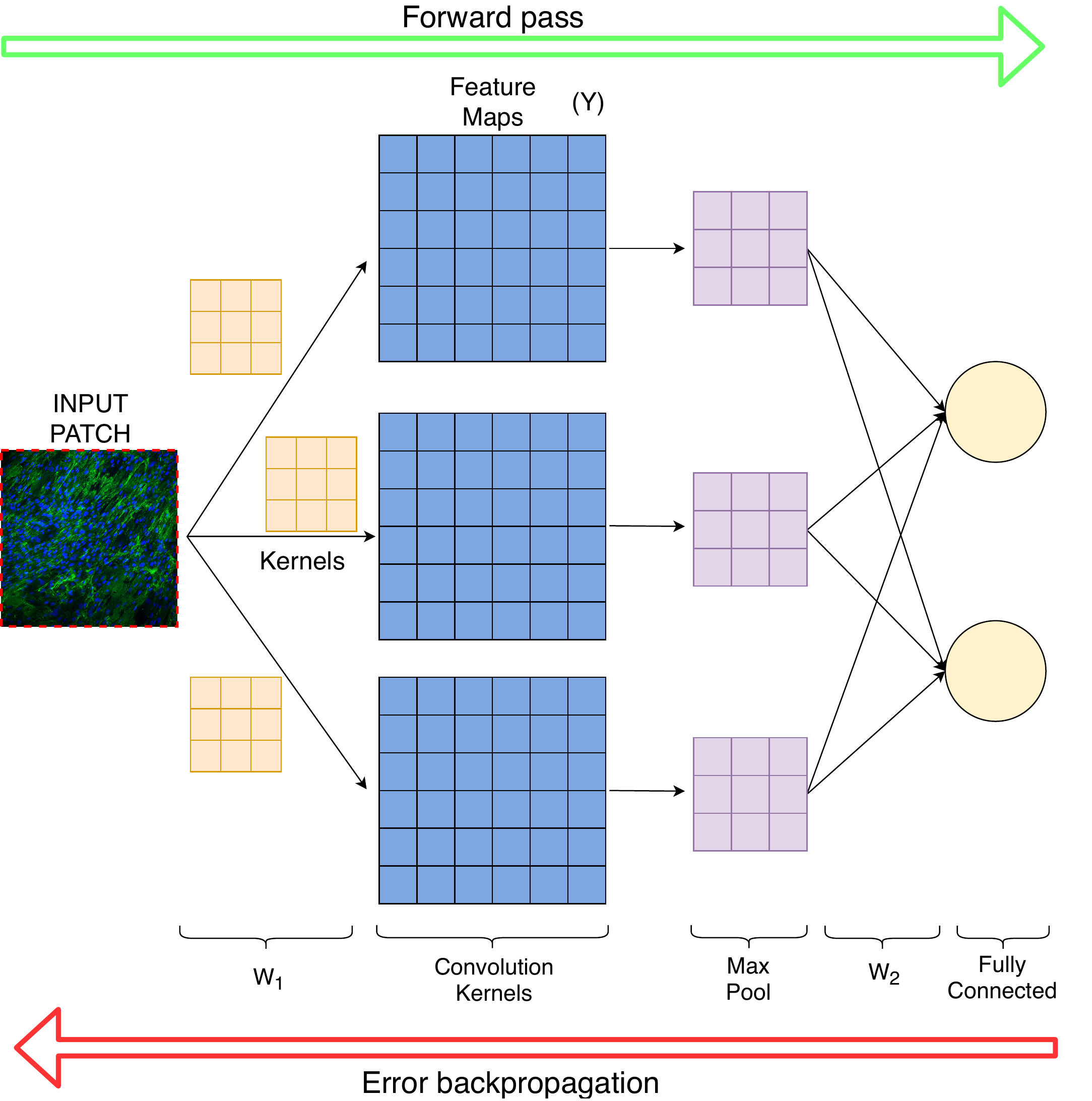}
  \caption{Process of training of a CNN using backpropagation. A forward pass through the CNN calculates the activations of neurons. The error between the output and the known ground-truth is sent back through the CNN to compute the gradient of the error with respect to the weights ($W_1$ and $W_2$ in this illustration). Then, gradient descent is used to update the weights.}
  \label{CNNTraining1}
\end{figure}

As depicted in Fig.~\ref{CNNTraining1}, a CNN learns comparing its outputs with the ones given in a labeled dataset. Specifically, we want to make the output $o_i$ and the target $t_i$ to be as similar as possible for each input instance $i$ = 1,...,$n$, where $n$ is the number of instances. More formally, we want to minimize the error 
\begin{equation}
E=\frac{1}{2} \sum_{i=1}^{n} ||o_i(W) - t_i||^2, \label{updatecnn1}
\end{equation}
with $W$ the weights in the CNN determining the output. The weights are initialized with random values and they are updated with a gradient descent rule as
\begin{equation}
W = W - \alpha\: \frac{\partial E}{\partial W}, \label{updatecnn2}
\end{equation}
where $\alpha$ is the learning rate determining the intensity of the gradient descent during training. The backpropagation algorithm~\cite{Hung:1993} is an efficient way to compute the gradient of the error with respect to the weights and it is the approach used in this work. 

For binary classification tasks, the available training data is split in two sets: a training set typically containing about 80\% of the data and a test set. A particular batch of training inputs is used at each iteration and the process is repeated until all the training inputs are used. The iteration over all the available training set is known as an epoch. After each epoch, the performance of the CNN is evaluated using a validation set including 10\% of the training data.  Finally, after training for a certain number of epochs, the CNN generalization capability is evaluated using the testing set, that was not used for training.

The evaluation with a validation and a testing set reduces overfitting, that is a major issue  happening when a CNN perfectly classifies the training data, but is unable to correctly classify unseen cases. Dropout is another mechanism to prevent overfitting~\cite{Srivastava:2014}. With this mechanism, some neurons are deactivated with a given probability. These elements do not participate in the activation of posterior neurons nor in the error correction, i.e., they are virtually removed from the CNN. The benefit is that the network becomes less sensitive to the specific weights.

%-----------------------------------------------------------------------------------------------------
%-----------------------------------------------------------------------------------------------------
%-----------------------------------------------------------------------------------------------------

\begin{figure}[t]
  \centering
  \includegraphics[scale=0.4]{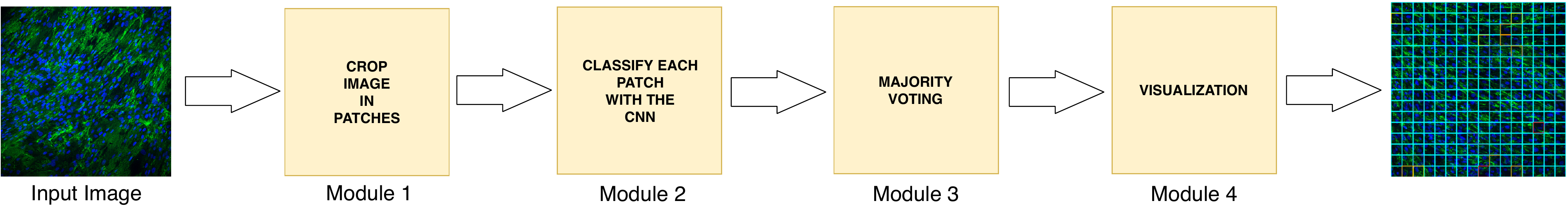}
  \caption{Overview of the proposed automatic diagnosis system using a majority voting on the individual patches decisions of the CNN model. The system also provides a detailed visualization of the diagnosis.}
  \label{fig:diagnosissystem}
\end{figure}

\begin{figure}[t]
  \centering
  \includegraphics[scale=0.65]{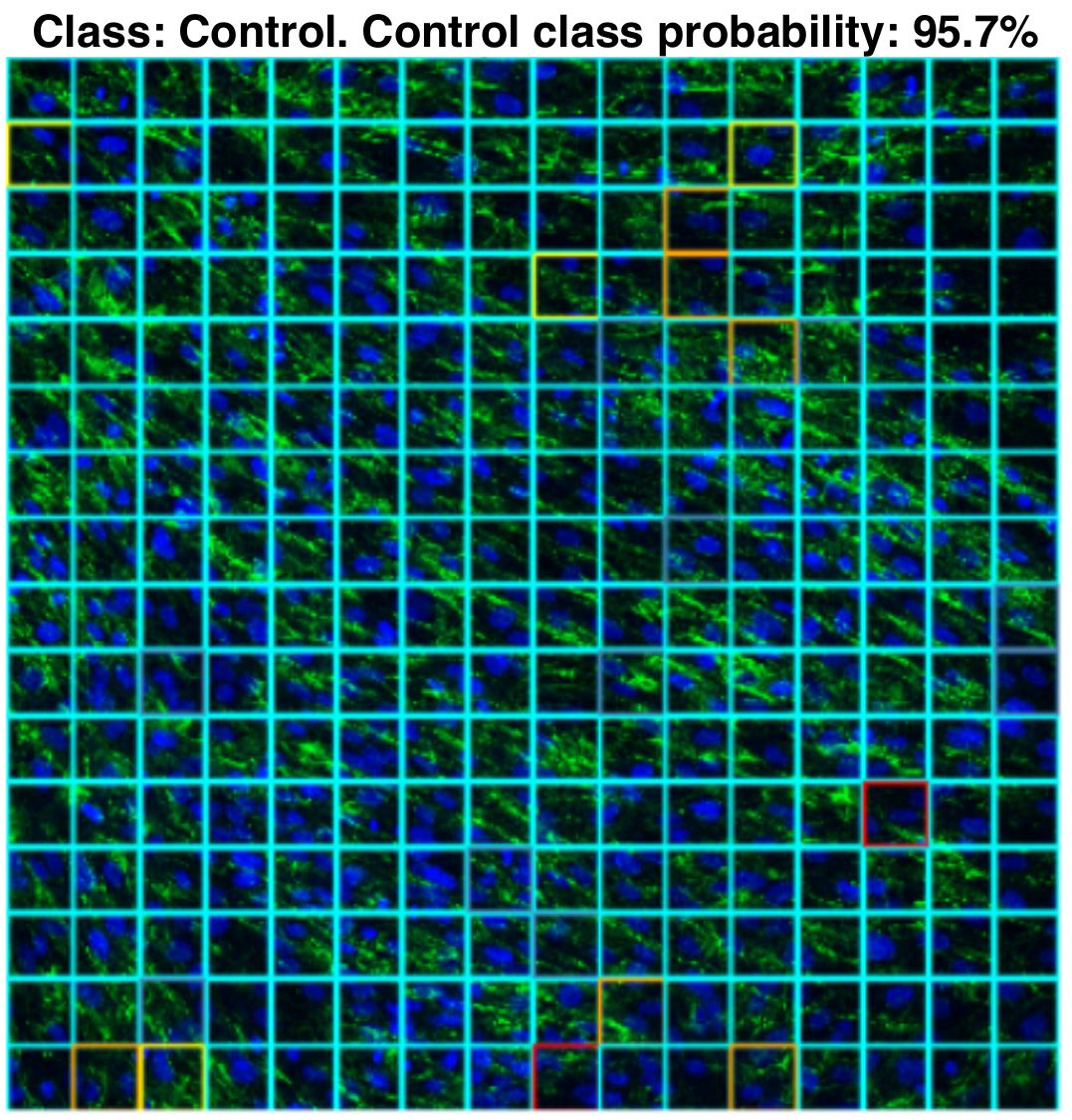}
  \caption{Visualization of the diagnosis of a given fibroblast culture image. Each patch of the image is colored according to its probability of belonging to the control class. The system also gives an overall score computed as the percentage of
  patches classified as control in the image.}
  \label{fig:CADvisualization}
\end{figure}

\section{The proposed system} \label{sec:system}

Figure~\ref{fig:diagnosissystem} provides a global view of the proposed CAD system. The system is divided in four modules. The first module receives a full image and splits it into non-overlapping patches of 64x64 pixels. The second module is formed by the CNN classification model, that receives the patches and outputs an independent prediction for each one of them. The third module receives the local decisions for each patch and takes a global decision using majority voting. The last module visualizes the input image, the decision for each patch represented in a color code, and the overall decision of the system (see Fig.~\ref{fig:CADvisualization}). Cyan is used to frame patches with more than 90\% probability of belonging to the control class, steel blue is used for patches with probability between 70\% and 90\%, yellow for patches with probability between 50\% and 70\%, orange for patches with probability between 30\% and 50\%, and finally red for patches with less than 30\% of probability of belonging to the control class. This color code offers the possibility of easily spotting suspicious areas in the image. The system also provides the overall decision on the image and a global score computed as the percentage of patches classified as control in the image. This score enables to track the evolution of a patient and, thus, it offers a tool to assess whether or not a particular treatment is effective.

The core component in the proposed approach is the CNN model used to classify the image patches. 
We experimentally identified a CNN architecture suitable for the addressed problem. In this experimental selection two concepts were considered: the size of the network (with small/large number of convolution kernels) and the abstraction level (with increasing/decreasing sizes of the subsequent layers of the CNN). Experimentally, small CNNs with decreasing number of features provided the best results. A key feature in such architectures, is that many low-level features are learned on the first layers and few high-level discriminating features of the images are generated in deeper layers. The proposed architecture is detailed in Table \ref{tab:DetailsCNNTable} and is similar to the one in~\cite{Krizhevsky:2012}, which has already been proven to be particularly adequate for image classification. However, our network is smaller since the classification task addressed here is simpler than the one addressed in~\cite{Krizhevsky:2012}.

\begin{table}[t]
\centering
\caption{The details of the CNN architecture proposed in this paper.}
   \label{tab:DetailsCNNTable}
   \begin{tabular}{C{1.5cm}*3{C{3cm}}C{2cm}}
     \hline
Layer & Type & Number of neurons (output layer)  & Kernel size & Stride \\
     \hline
1 & Convolution & 64x64x128 & 3x3 & 1 \\ 
2 & Max Pooling & 32x32x128 & 2x2 & 2 \\ 
3 & Convolution & 32x32x64 & 3x3 & 1 \\ 
4 & Max Pooling & 16x16x64 & 2x2 & 2 \\ 
5 & Convolution & 16x16x32 & 3x3 & 1 \\ 
6 & Max Pooling & 8x8x32 & 2x2 & 2 \\ 
7 & Fully Connected & 150 & - & - \\
8 & Fully Connected & 2 & - & - \\
\hline
\end{tabular}
\end{table}

\begin{figure*}[t]
  \includegraphics[width=\textwidth,height=2.5cm]{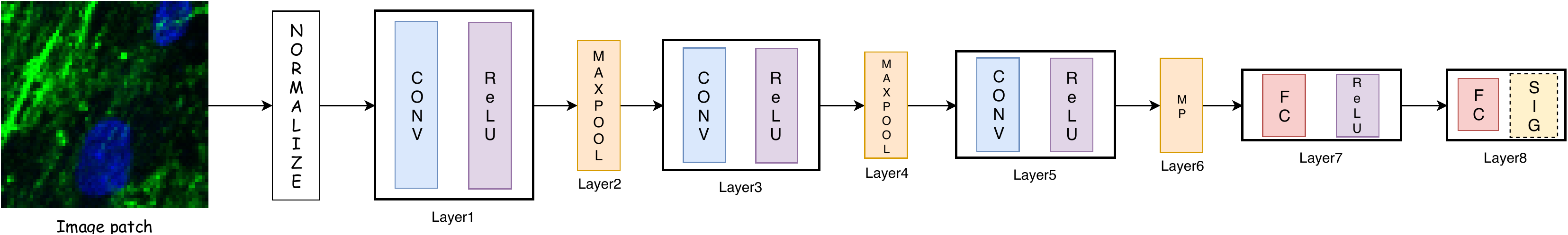}
  \caption{The proposed CNN model architecture} \label{fig:architecture}
\end{figure*}

Figure~\ref{fig:architecture} provides a detailed overview of the CNN architecture proposed in this paper. The input image patch is normalized in order to have zero mean and unit variance. Normalization improves the gradient descent process and avoids its premature convergence. The normalized image patch is passed through three convolutional layers (layers 1, 3, and 5) which include the application of ReLU activation functions after the kernel computation. The first convolution layer defines 128 feature maps of size 64x64, the second one defines 64 feature maps of size 32x32, and the last one defines 32 feature maps of size 16x16. The reduction in the size of the feature maps is obtained with  max-pooling layers (layers 2, 4, and 6) with a stride of 2 following the convolution layers. After the feature generation layers, the classification is implemented with two fully-connected layers (layers~7 and~8). The first one has 150 neurons, also followed by a ReLU activation function trained with a dropout mechanism with probability 0.5. The second fully connected layer has 2 neurons whose output is truncated into a single binary output by a sigmoid activation function to provide the final classification.

%-----------------------------------------------------------------------------------------------------
%-----------------------------------------------------------------------------------------------------
%-----------------------------------------------------------------------------------------------------

\section{Data acquisition and augmentation} \label{sec:setup}

Samples from the forearm were obtained from possible patients with as well as from aged-matched controls. Primary fibroblasts cultures were established using standard procedures~\cite{Jimenez-Mallebrera:2006}. Confluent fibroblasts (patient and control in parallel) were treated with 25 $\mu$g/mL of L-ascorbic acid phosphate magnesium (Wako Chemicals GmbH, Neuss, Germany) for 24 hours. After that time, cells were fixed with 4\% paraformaldehyde in phosphate-buffered saline solution. Collagen VI was detected by indirect immunofluorescence using a monoclonal antibody (MAB1944, Merck, Germany) as previously described in~\cite{Jimenez-Mallebrera:2006} and fibroblast nuclei were stained using 4,6-diamidino-2-phenylindole  (Sigma Chemical, St. Louis, USA).  

The images were acquired with a Leica TCS SP8 X White Light Laser confocal microscope with hybrid spectral detectors (Leica Microsystems, Wetzlar, Germany). The confocal images were acquired using a HC x PL APO 20x/0.75 dry objective and with the confocal pinhole set to 1 Airy unit. Collagen VI was excited with an argon laser (488 nm) and detected in the 500-560 nm and nuclei were excited with a blue diode laser (405 nm) and detected in the 420–460 nm. Appropriate negative controls were used to adjust confocal settings to avoid non-specific fluorescence artifacts. The detector gain and offset values were adjusted to use the entire dynamic rate of the detector (12 bits) and to avoid oversaturated voxels. Sequential acquisition settings were used to avoid inter-channel cross-talk. Ten sections of each sample were acquired every 1.5 $\mu$m along the focal axis (Z-stack) and combined into an integrated intensity projection to form a single image. 

Since the data acquisition procedure is complex and we are dealing with a rare disease, the available data is limited. Thus, we propose a data augmentation scheme inspired in the one described in~\cite{Krizhevsky:2012} to generate enough inputs to train the CNN.

The data is augmented by splitting the given full size images into small, non-overlapping patches. Each patch is used as an independent input to train the CNN model. In our approach, the patch is of size 64x64 pixels since it has been shown to be a particularly relevant window size for CNN classification models on similar images~\cite{Spanhol:2016}. In our case, the patches capture the main features of the input images, i.e., a significant portion of the collagen network and several nuclei or parts of them.

Since our input images are of 1024x1024 pixels, 256 patches per image are generated. Each patch is further transformed to get even more variations of the data: They are rotated clockwise by 90, 180, 270 and 360 degrees and every rotated patch is flipped horizontally. Thus, eight different variations of each original patch can be obtained and, consequently, each input image generates 2048 training inputs for the CNN. Since initially we have 276 images, with this data augmentation process, the training  and testing sets include, respectively, 56320 and 14336 patches, without taking into account the rotated and mirrored patches.

%14233
%-25

\begin{figure}[t]
  \centering
  \includegraphics[scale=0.75]{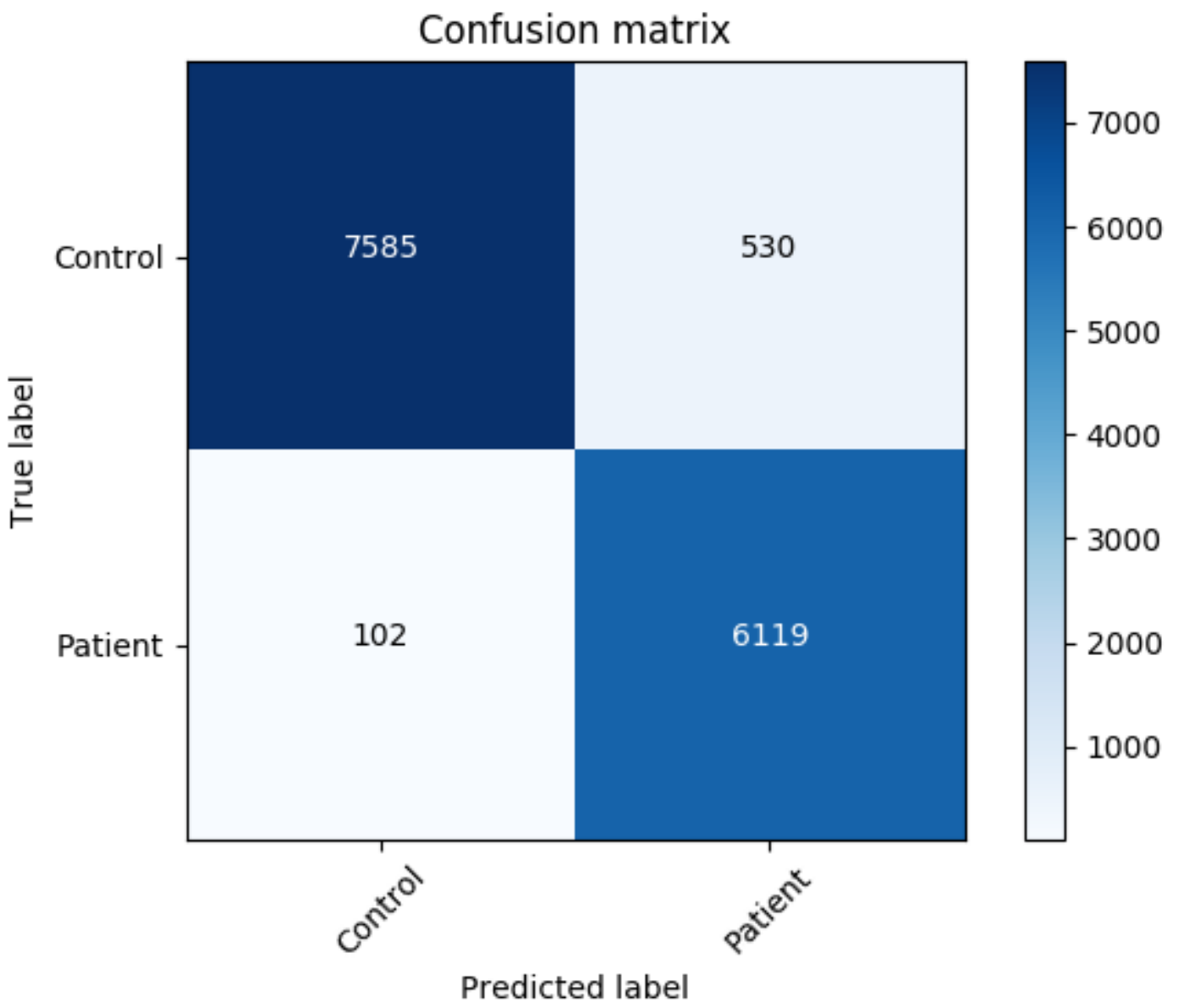}
  \caption{Confusion matrix of the test set for the model trained with the 64x64 image patches.}
  \label{fig:ConfusionMatrixFig}
\end{figure}

%-----------------------------------------------------------------------------------------------------
%-----------------------------------------------------------------------------------------------------
%-----------------------------------------------------------------------------------------------------

\section{Results and discussion} \label{sec:results}

The proposed system was implemented using the Python programming language and the Keras 2.2 library with TensorFlow 0.19 backend and it was executed on a workstation with a Intel Core i7-7700HQ processor and 16 GB of RAM. All the processing was performed in a NVIDIA GeForce GTX 1050 GPU. The batch size is set to 32 and the training is executed while 
the performance of the classification of the test set improves. On average, each epoch took about 14 minutes to complete and less than 10 epochs are necessary to converge. For training the neural network we use the Adam~\cite{Kingma:2014} optimizer, with a learning rate of $\alpha=0.01$, an exponential decay rate for the first moment estimates of 
$\beta_1 = 0.9$, an exponential decay rate for the second moment estimates of $\beta_2 = 0.999$, and $\epsilon = 10^{-8}$ to prevent any division by zero.

We present results at two different levels. The first level is the per-patch performance of the CNN model and the second is the performance of the proposed system on full images, by using majority voting. In this way, we evaluate how the model performs using only local features (patches) and how it improves when integrating local results to take a global decision (full images).

Figure~\ref{fig:ConfusionMatrixFig} gives the confusion matrix of the system diagnosing on 64x64 patches on the test set. In this confusion matrix, the amount of patches correctly and incorrectly classified  as belonging to the control class are in the first row. From a total of 8115 control patches in the test set, 7585 are true negatives ($t_n$), i.e., the inputs correctly classified as control and 530 are false positives ($f_p$), i.e., the inputs incorrectly classified as patient. The classification of patients is given in the second row, where, 102 are false negatives ($f_n$), i.e., the inputs incorrectly classified as control, and 6119 are true positives ($t_p$), i.e., the inputs correctly classified as patient. In the majority voting system the accuracy is perfect and, thus, the confusion matrix is trivial and not given here.

\begin{table}[t]
\centering
\caption{The performance of the proposed systems on the test set.}
   \label{tab:PerformanceTable}
   \begin{tabular}{C{2cm}C{2cm}C{1.5cm}C{1.5cm}C{1.5cm}C{1.5cm}}
     \hline
     System & Size      &   $A$  &  $P$ &  $R$ & $F_1$ \\ 
     \hline
     Patch  & 64x64     &  0.95  & 0.92 & 0.98 &  0.95 \\ 
     Image  & 1024x1024 &  1     & 1   &  1   &   1    \\ 
\hline
\end{tabular}
\end{table}

Table~\ref{tab:PerformanceTable} gives the performance evaluation metrics for each of the analyzed levels. 
The accuracy, $A$, precision, $P$, recall, $R$, and $F_1$ score are used to assess the classification performance. The accuracy,
\begin{equation}
A = \frac{t_p + t_n}{t_p + f_p + t_n + f_n}, \label{eq:accuracy}
\end{equation}
refers to the correct classification rate, defined as the ratio of correctly classified cases with respect to the total number of cases. The precision, 
\begin{equation}
P = \frac{t_p}{t_p + f_p}, \label{eq:precision}  
\end{equation}
is the ratio of inputs correctly classified as patient with respect to the total number of inputs (correctly or incorrectly) classified a patient.
The recall, 
\begin{equation}
R = \frac{t_p}{t_p + f_n}, \label{eq:recall} 
\end{equation}
gives the ratio of samples correctly classified as patient with respect to total inputs that are actually in the patient class. Finally, the $F_1$ metric 
\begin{equation}
F_1 = \frac{2\:P\:R}{P + R}, \label{eq:f1} 
\end{equation}
is the weighted harmonic mean of precision and recall. 

At the level of patch classification, the system achieves an accuracy of 0.95, a precision of 0.92, and a recall of 0.98. Consequently, the proposed system attains a $F_1$ score of 0.95. At the level of images, the system achieves perfect classification results.

\begin{figure}[t]
  \centering
  \includegraphics[scale=0.75]{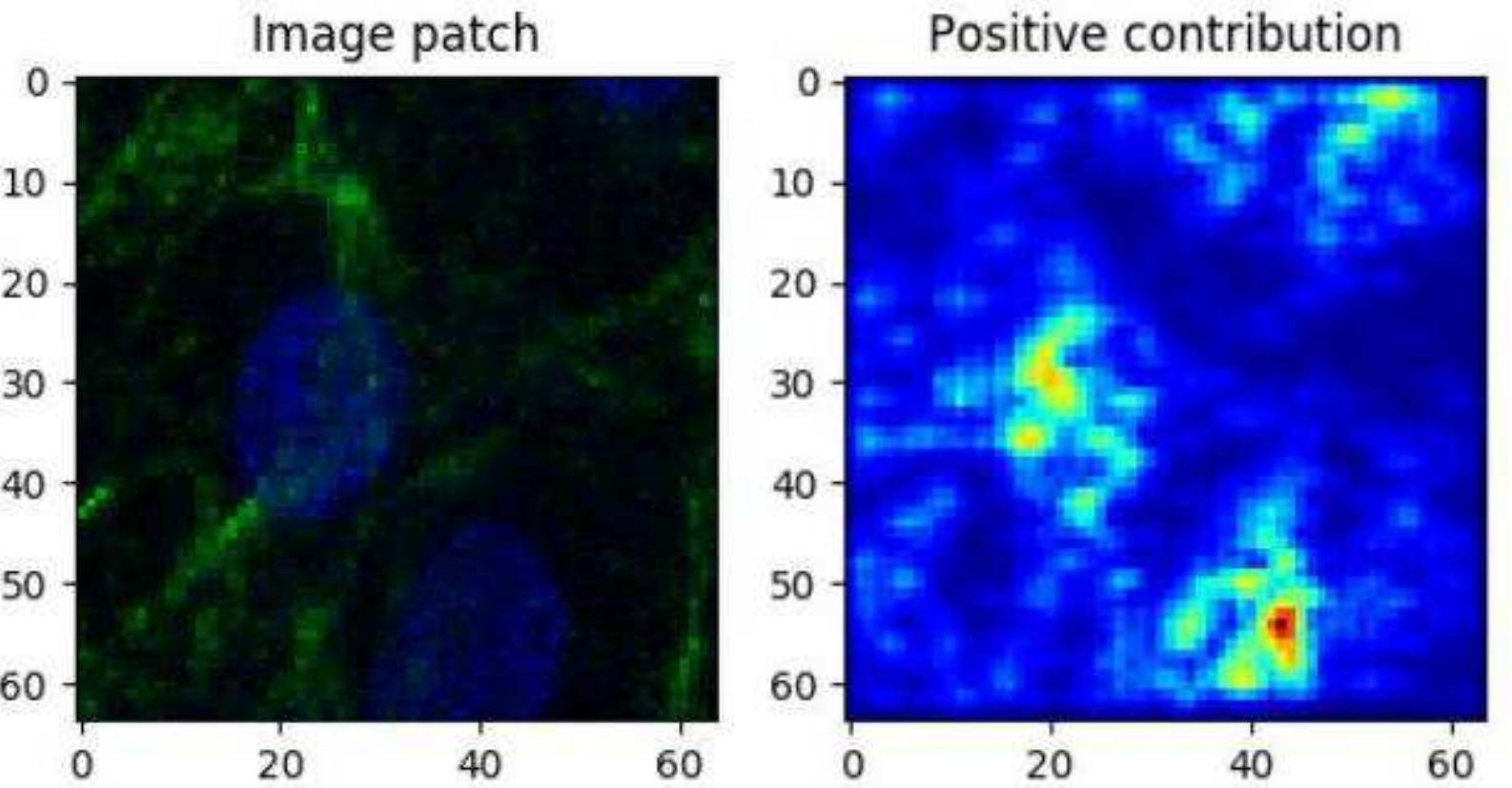}
  \caption{Left: Visualization of an image patch. Right: The saliency map with respect to the control class (in red the pixels with higher contribution to the classification).}
  \label{fig:interpretability1}
\end{figure}

\begin{figure}[t]
  \centering
  \includegraphics[scale=0.4]{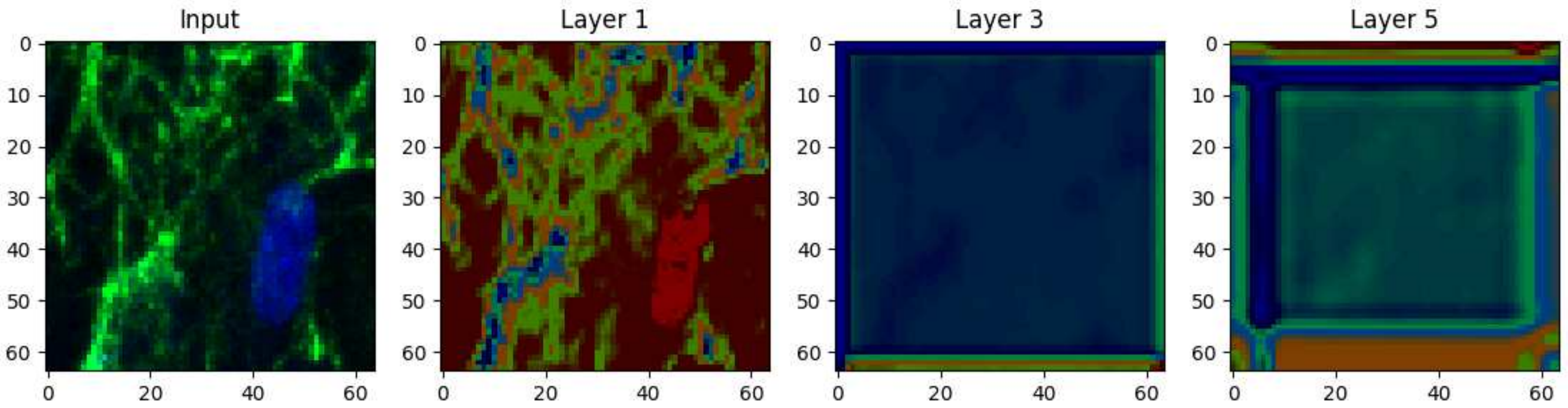}
  \caption{Visualization of the activation of each convolutional layer with respect to the patient class.}
  \label{fig:interpretability2}
\end{figure}

CNNs can provide very accurate results, but they are often seen as black boxes since it is hard to find out what are the features and the classification criteria learned from the data. However, recent works try to interpret the CNNs after the training phase~\cite{Simonyan:2013}. Following this approach, we provide a visualization of two relevant parts of the CNN.
First, Fig.~\ref{fig:interpretability1} shows the contribution of each pixel in the image to the classification as control class. The network focus on the fibroblast nuclei and on the regions without collagen~VI. Note that the intensity on the later is lower, but their extension is larger and, thus its actual contribution to the classification is also significant.
In second term, Fig.~\ref{fig:interpretability2} shows the saliency maps of the convolutional layers with respect to the patient class, where the saliency is the gradient of the output with respect to the corresponding convolutional layer. At early layers, the network learns to detect low level features, such as the presence of fibroblasts or the lack of collagen VI, confirming the previous analysis. In posterior layers, the features become more abstract and, thus, more difficult to relate with elements in the input images.

%-----------------------------------------------------------------------------------------------------
%-----------------------------------------------------------------------------------------------------
%-----------------------------------------------------------------------------------------------------

\section{Conclusions} \label{sec:conclusions} 

This paper describes a system for the computer aided diagnosis of muscular dystrophies caused by deficiencies in the structure of collagen~VI. The proposed system relies on a deep convolutional neural network and on a data augmentation scheme to handle the problem of lack of data typical of rare diseases. The proposed system is capable of achieving perfect results in the diagnosis task in the available testing set. This provides a solid tool to automatically and accurately track the effect of potential therapies for the recovery of collagen~VI in patients. Furthermore, we visualized the contribution of each pixel to the classification as well as the features learned by the CNN in each convolutional layer by means of saliency maps. This identifies the primary features discovered by the CNN. Our current research endeavours focus on extending this analysis to clarify the abstract features and the classification criteria identified by the proposed CNN. Moreover, we are studying the possible application of image-based computer aided diagnosis procedures to other rare diseases related to deficiencies in collagen~VI or other extra-cellular matrix proteins.

%-----------------------------------------------------------------------------------------------------
%-----------------------------------------------------------------------------------------------------
%-----------------------------------------------------------------------------------------------------

\section*{Ethics Statement}
This study was carried out in accordance with the recommendations of the Fundación Sant Joan de Déu Ethics Committee. Written informed consent was obtained from patients and/or their parents or guardians in accordance with the Declaration of Helsinki. The protocol was approved by the Fundación Sant Joan de Déu Ethics Committee. 

%-----------------------------------------------------------------------------------------------------
%-----------------------------------------------------------------------------------------------------
%-----------------------------------------------------------------------------------------------------

\section*{Acknowledgements}

Adrián Bazaga was supported by a JAE-intro scholarship granted by the {\em Spanish Council of Scientific Research}. Josep M. Porta is funded by the Spanish Ministry of Economy and Competitiveness under project DPI2017-88282-P.

Cecilia Jiménez-Mallebrera and Carmen Badosa are funded by the Health Institute 'Carlos III' (ISCIII, Spain) and the European Regional Development Fund (ERDF/FEDER), 'A way of making Europe', grants references PI16/00579, CP09/00011 and Fundación Noelia.

%-----------------------------------------------------------------------------------------------------
%-----------------------------------------------------------------------------------------------------
%-----------------------------------------------------------------------------------------------------

\section*{Author Contributions}

Adrián Bazaga -- Conceptualization; Investigation; Formal analysis; Methodology; Software; Validation; Visualization; Writing - original draft; Writing - review \& editing.

Mònica Roldán -- Data curation; Resources; Writing - original draft; Writing - review \& editing.

Carmen Badosa -- Data curation; Writing - review \& editing.

Cecilia Jiménez-Mallebrera -- Data curation; Resources; Writing - original draft; Writing - review \& editing.

Josep M. Porta -- Conceptualization; Formal analysis; Methodology; Supervision; Funding acquisition; Writing - original draft; Writing - review \& editing.

%-----------------------------------------------------------------------------------------------------
%-----------------------------------------------------------------------------------------------------
%-----------------------------------------------------------------------------------------------------


\begin{thebibliography}{10}
\expandafter\ifx\csname url\endcsname\relax
  \def\url#1{\texttt{#1}}\fi
\expandafter\ifx\csname urlprefix\endcsname\relax\def\urlprefix{URL }\fi
\expandafter\ifx\csname href\endcsname\relax
  \def\href#1#2{#2} \def\path#1{#1}\fi

\bibitem{Nadeau:2009}
A.~Nadeau, M.~Kinali, M.~Main, C.~Jimenez-Mallebrera, A.~Aloysius, E.~Clement,
  B.~North, A.~Y. Manzur, S.~A. Robb, E.~Mercuri, F.~Muntoni,
  \href{https://doi.org/10.1212/WNL.0b013e3181aae851}{Natural history of
  {Ullrich} congenital muscular dystrophy}, Neurology 73~(1) (2009) 25--31.
\newline\urlprefix\url{https://doi.org/10.1212/WNL.0b013e3181aae851}

\bibitem{Lamand:2018}
S.~R. Lamand{\'{e}}, J.~F. Bateman,
  \href{https://doi.org/10.1016/j.matbio.2017.12.008}{Collagen {VI} disorders:
  Insights on form and function in the extracellular matrix and beyond}, Matrix
  Biology 71-72 (2018) 348--367.
\newline\urlprefix\url{https://doi.org/10.1016/j.matbio.2017.12.008}

\bibitem{Jimenez-Mallebrera:2006}
C.~Jimenez-Mallebrera, M.~Maioli, J.~Kim, S.~Brown, L.~Feng, A.~Lampe,
  K.~Bushby, D.~Hicks, K.~Flanigan, C.~Bonnemann, C.~Sewry, F.~Muntoni,
  \href{https://doi.org/10.1016/j.nmd.2006.07.015}{A comparative analysis of
  collagen vi production in muscle, skin and fibroblasts from 14 {Ullrich}
  congenital muscular dystrophy patients with dominant and recessive col6a
  mutations}, Neuromuscular Disorders 16~(9) (2006) 571--582.
\newline\urlprefix\url{https://doi.org/10.1016/j.nmd.2006.07.015}

\bibitem{Anthony:2014}
K.~Anthony, V.~Arechavala-Gomeza, L.~E. Taylor, A.~Vulin, Y.~Kaminoh,
  S.~Torelli, L.~Feng, N.~Janghra, G.~Bonne, M.~Beuvin, R.~Barresi,
  M.~Henderson, S.~Laval, A.~Lourbakos, G.~Campion, V.~Straub, T.~Voit, C.~A.
  Sewry, J.~E. Morgan, K.~M. Flanigan, F.~Muntoni,
  \href{https://doi.org/10.1212/WNL.0000000000001025}{Dystrophin
  quantification}, Neurology 83~(22) (2014) 2062--2069.
\newline\urlprefix\url{https://doi.org/10.1212/WNL.0000000000001025}

\bibitem{Krizhevsky:2012}
A.~Krizhevsky, I.~Sutskever, G.~E. Hinton,
  \href{http://dl.acm.org/citation.cfm?id=2999134.2999257}{Imagenet
  classification with deep convolutional neural networks}, in: Proceedings of
  the 25th International Conference on Neural Information Processing Systems -
  Volume 1, 2012, pp. 1097--1105.
\newline\urlprefix\url{http://dl.acm.org/citation.cfm?id=2999134.2999257}

\bibitem{Schilham:2006}
A.~M. Schilham, B.~van Ginneken, M.~Loog,
  \href{https://doi.org/10.1016/j.media.2005.09.003}{A computer-aided diagnosis
  system for detection of lung nodules in chest radiographs with an evaluation
  on a public database}, Medical Image Analysis 10~(2) (2006) 247--258.
\newline\urlprefix\url{https://doi.org/10.1016/j.media.2005.09.003}

\bibitem{Adankon:2012}
M.~M. Adankon, J.~Dansereau, H.~Labelle, F.~Cheriet,
  \href{https://doi.org/10.1016/j.artmed.2012.07.002}{Non invasive
  classification system of scoliosis curve types using least-squares support
  vector machines}, Artificial Intelligence in Medicine 56~(2) (2012) 99--107.
\newline\urlprefix\url{https://doi.org/10.1016/j.artmed.2012.07.002}

\bibitem{Ahonen:2006}
T.~Ahonen, A.~Hadid, M.~Pietikainen,
  \href{https://doi.org/10.1109/TPAMI.2006.244}{Face description with local
  binary patterns: Application to face recognition}, IEEE Transactions on
  Pattern Analysis and Machine Intelligence 28~(12) (2006) 2037--2041.
\newline\urlprefix\url{https://doi.org/10.1109/TPAMI.2006.244}

\bibitem{Guo:2010}
Z.~Guo, L.~Zhang, D.~Zhang, \href{https://doi.org/10.1109/TIP.2010.2044957}{A
  completed modeling of local binary pattern operator for texture
  classification}, IEEE Transactions on Image Processing 19~(6) (2010)
  1657--1663.
\newline\urlprefix\url{https://doi.org/10.1109/TIP.2010.2044957}

\bibitem{Soh:1999}
L.-K. Soh, C.~Tsatsoulis, \href{https://doi.org/10.1109/36.752194}{Texture
  analysis of {SAR} sea ice imagery using gray level co-occurrence matrices},
  {IEEE} Transactions on Geoscience and Remote Sensing 37~(2) (1999) 780--795.
\newline\urlprefix\url{https://doi.org/10.1109/36.752194}

\bibitem{Khatami:2017}
A.~Khatami, A.~Khosravi, T.~Nguyen, C.~P. Lim, S.~Nahavandi,
  \href{https://doi.org/10.1016/j.eswa.2017.05.073}{Medical image analysis
  using wavelet transform and deep belief networks}, Expert Systems with
  Applications 86 (2017) 190--198.
\newline\urlprefix\url{https://doi.org/10.1016/j.eswa.2017.05.073}

\bibitem{Sun:2015}
M.~Sun, A.~B. Bloom, M.~H. Zaman,
  \href{https://doi.org/10.1371/journal.pone.0131814}{Rapid quantification of
  {3D} collagen fiber alignment and fiber intersection correlations with high
  sensitivity}, {PLOS} {ONE} 10~(7) (2015) e0131814.
\newline\urlprefix\url{https://doi.org/10.1371/journal.pone.0131814}

\bibitem{Subashini:2016}
M.~M. Subashini, S.~K. Sahoo, V.~Sunil, S.~Easwaran,
  \href{https://doi.org/10.1016/j.eswa.2015.08.036}{A non-invasive methodology
  for the grade identification of astrocytoma using image processing and
  artificial intelligence techniques}, Expert Systems with Applications 43
  (2016) 186--196.
\newline\urlprefix\url{https://doi.org/10.1016/j.eswa.2015.08.036}

\bibitem{Celebi:2007}
M.~E. Celebi, H.~A. Kingravi, B.~Uddin, H.~Iyatomi, Y.~A. Aslandogan, W.~V.
  Stoecker, R.~H. Moss,
  \href{https://doi.org/10.1016/j.compmedimag.2007.01.003}{A methodological
  approach to the classification of dermoscopy images}, Computerized Medical
  Imaging and Graphics 31~(6) (2007) 362--373.
\newline\urlprefix\url{https://doi.org/10.1016/j.compmedimag.2007.01.003}

\bibitem{Nahar:2012}
J.~Nahar, T.~Imam, K.~S. Tickle, A.~S. Ali, Y.-P.~P. Chen,
  \href{https://doi.org/10.1016/j.eswa.2012.04.045}{Computational intelligence
  for microarray data and biomedical image analysis for the early diagnosis of
  breast cancer}, Expert Systems with Applications 39~(16) (2012) 12371--12377.
\newline\urlprefix\url{https://doi.org/10.1016/j.eswa.2012.04.045}

\bibitem{Shabut:2018}
A.~M. Shabut, M.~H. Tania, K.~T. Lwin, B.~A. Evans, N.~A. Yusof, K.~J.
  Abu-Hassan, M.~Hossain, \href{https://doi.org/10.1016/j.eswa.2018.07.014}{An
  intelligent mobile-enabled expert system for tuberculosis disease diagnosis
  in real time}, Expert Systems with Applications 114 (2018) 65--77.
\newline\urlprefix\url{https://doi.org/10.1016/j.eswa.2018.07.014}

\bibitem{Venkatesan:2015}
R.~Venkatesan, P.~S. Chandakkar, B.~Li,
  \href{https://doi.org/10.1109/ICCV.2015.299}{Simpler non-parametric methods
  provide as good or better results to multiple-instance learning}, in: IEEE
  International Conference on Computer Vision (ICCV), 2015, pp. 2605--2613.
\newline\urlprefix\url{https://doi.org/10.1109/ICCV.2015.299}

\bibitem{Menegola:2016}
A.~Menegola, M.~Fornaciali, R.~Pires, S.~E.~F. de~Avila, E.~Valle,
  \href{http://arxiv.org/abs/1609.01228}{Towards automated melanoma screening:
  Exploring transfer learning schemes}, CoRR abs/1609.01228.
\newblock \href {http://arxiv.org/abs/1609.01228} {\path{arXiv:1609.01228}}.
\newline\urlprefix\url{http://arxiv.org/abs/1609.01228}

\bibitem{Vinicius:2018}
M.~V. dos Santos~Ferreira, A.~O. de~Carvalho~Filho, A.~D. de~Sousa, A.~C.
  Silva, M.~Gattass,
  \href{https://doi.org/10.1016/j.eswa.2018.06.010}{Convolutional neural
  network and texture descriptor-based automatic detection and diagnosis of
  glaucoma}, Expert Systems with Applications 110 (2018) 250--263.
\newline\urlprefix\url{https://doi.org/10.1016/j.eswa.2018.06.010}

\bibitem{Sudharshan:2019}
P.~Sudharshan, C.~Petitjean, F.~Spanhol, L.~E. Oliveira, L.~Heutte, P.~Honeine,
  \href{https://doi.org/10.1016/j.eswa.2018.09.049}{Multiple instance learning
  for histopathological breast cancer image classification}, Expert Systems
  with Applications 117 (2019) 103--111.
\newline\urlprefix\url{https://doi.org/10.1016/j.eswa.2018.09.049}

\bibitem{Lecun:1998}
Y.~Lecun, L.~Bottou, Y.~Bengio, P.~Haffner,
  \href{https://doi.org/10.1109/5.726791}{Gradient-based learning applied to
  document recognition}, Proceedings of the {IEEE} 86~(11) (1998) 2278--2324.
\newline\urlprefix\url{https://doi.org/10.1109/5.726791}

\bibitem{Simonyan:2014}
K.~Simonyan, A.~Zisserman, \href{https://arxiv.org/abs/1409.1556}{Very deep
  convolutional networks for large-scale image recognition}, CoRR
  abs/1409.1556.
\newblock \href {http://arxiv.org/abs/1409.1556} {\path{arXiv:1409.1556}}.
\newline\urlprefix\url{https://arxiv.org/abs/1409.1556}

\bibitem{Goodfellow:2016}
I.~Goodfellow, Y.~Bengio, A.~Courville,
  \href{https://www.deeplearningbook.org}{Deep Learning}, MIT Press, 2016.
\newline\urlprefix\url{https://www.deeplearningbook.org}

\bibitem{Hung:1993}
S.~Hung, H.~Adeli, \href{https://doi.org/10.1016/0925-2312(93)90042-2}{Parallel
  backpropagation learning algorithms on {CRAY Y-MP8/864} supercomputer},
  Neurocomputing 5~(6) (1993) 287--302.
\newline\urlprefix\url{https://doi.org/10.1016/0925-2312(93)90042-2}

\bibitem{Srivastava:2014}
N.~Srivastava, G.~Hinton, A.~Krizhevsky, I.~Sutskever, R.~Salakhutdinov,
  \href{http://jmlr.org/papers/v15/srivastava14a.html}{Dropout: A simple way to
  prevent neural networks from overfitting}, Journal of Machine Learning
  Research 15 (2014) 1929--1958.
\newline\urlprefix\url{http://jmlr.org/papers/v15/srivastava14a.html}

\bibitem{Spanhol:2016}
F.~A. Spanhol, L.~S. Oliveira, C.~Petitjean, L.~Heutte,
  \href{https://doi.org/10.1109/IJCNN.2016.7727519}{Breast cancer
  histopathological image classification using convolutional neural networks},
  in: International Joint Conference on Neural Networks (IJCNN), 2016, pp.
  2560--2567.
\newline\urlprefix\url{https://doi.org/10.1109/IJCNN.2016.7727519}

\bibitem{Kingma:2014}
D.~P. Kingma, J.~Ba, \href{http://arxiv.org/abs/1412.6980}{Adam: {A} method for
  stochastic optimization}, CoRR abs/1412.6980.
\newblock \href {http://arxiv.org/abs/1412.6980} {\path{arXiv:1412.6980}}.
\newline\urlprefix\url{http://arxiv.org/abs/1412.6980}

\bibitem{Simonyan:2013}
K.~Simonyan, A.~Vedaldi, A.~Zisserman,
  \href{http://arxiv.org/abs/1312.6034}{Deep inside convolutional networks:
  Visualising image classification models and saliency maps}, CoRR
  abs/1312.6034.
\newblock \href {http://arxiv.org/abs/1312.6034} {\path{arXiv:1312.6034}}.
\newline\urlprefix\url{http://arxiv.org/abs/1312.6034}

\end{thebibliography}
\end{document}